\documentclass{article}

\usepackage[numbers]{natbib}
\usepackage[final]{neurips_2022}

\usepackage[utf8]{inputenc} 
\usepackage[T1]{fontenc}    
\usepackage{url}            
\usepackage{booktabs}       
\usepackage{amsfonts}       
\usepackage{nicefrac}       
\usepackage{microtype}      
\usepackage{xcolor}         

\usepackage{amsmath}
\usepackage{amssymb}
\usepackage{graphicx}
\usepackage{booktabs}
\usepackage{xfrac}
\usepackage{color}
\usepackage{colortbl}
\usepackage{sidecap}
\usepackage{enumitem}
\usepackage{adjustbox}
\usepackage{romannum}
\usepackage{multirow}
\usepackage{wrapfig}
\usepackage{caption}
\usepackage{subcaption}
\usepackage{xspace}
\usepackage{hyperref}       
\definecolor{grey}{RGB}{128,138,135}
\definecolor{darkgrey}{RGB}{96,96,96}

\newcommand{\txt}[1]{{\texttt{#1}}}
\def\etal{\emph{et al.}\xspace}
\def\ie{\emph{i.e.}\xspace}
\def\eg{\emph{e.g.}\xspace}

\newcommand\extrafootertext[1]{%
    \bgroup
    \renewcommand\thefootnote{\fnsymbol{footnote}}%
    \renewcommand\thempfootnote{\fnsymbol{mpfootnote}}%
    \footnotetext[0]{#1}%
    \egroup
}

\definecolor{lavender}{HTML}{E5E2FB}
\definecolor{lightblue}{HTML}{CEEBF9}
\definecolor{lightyellow}{HTML}{F7D9AE}
\definecolor{lightred}{HTML}{EEDCDB}
\definecolor{green}{HTML}{3BCB41}
\definecolor{darkgreen}{HTML}{156C09}
\definecolor{purple}{HTML}{9903F0}

\usepackage{titlesec}
\usepackage{fontawesome}
\titlespacing\section{0pt}{8pt plus 4pt minus 2pt}{0pt plus 2pt minus 2pt}
\titlespacing\subsection{0pt}{8pt plus 4pt minus 2pt}{0pt plus 2pt minus 2pt}
\titlespacing\subsubsection{0pt}{8pt plus 4pt minus 2pt}{0pt plus 2pt minus 2pt}
\title{Bridging the Gap between Object and Image-level Representations for Open-Vocabulary Detection}

\author{%
  Hanoona Rasheed\textsuperscript{\textnormal{1,*}}, 
  Muhammad Maaz\textsuperscript{\textnormal{1,*}},
   Muhammad Uzair Khattak\textsuperscript{\textnormal{1}},\quad \\ 
  \textbf{Salman Khan}\textsuperscript{\textnormal{1,2}},
  \textbf{Fahad Shahbaz Khan}\textsuperscript{\textnormal{1,3}} \\
  \textsuperscript{1}Mohamed bin Zayed University of AI, UAE \\ 
  \textsuperscript{2}Australian National University, Australia \quad \textsuperscript{3}Linköping University, Sweden
    }

\begin{document}

\maketitle

\begin{abstract}
Existing open-vocabulary object detectors typically enlarge their vocabulary sizes by leveraging different forms of weak supervision. This helps generalize to novel objects at inference. Two popular forms of weak-supervision used in open-vocabulary detection (OVD) include pretrained CLIP model and image-level supervision. We note that both these modes of supervision are \emph{not} optimally aligned for the detection task: CLIP is trained with image-text pairs and lacks precise localization of objects while the image-level supervision has been used with heuristics that do not accurately specify local object regions. In this work, we propose to address this problem by performing object-centric alignment  of the language embeddings from the CLIP model. Furthermore, we visually ground the objects with only image-level supervision using a pseudo-labeling process that provides high-quality object proposals and helps expand the vocabulary during training. We establish a bridge between the above two object-alignment strategies via a novel weight transfer function that aggregates their complimentary strengths. In essence, the proposed model seeks to minimize the gap between object and image-centric representations in the OVD setting. On the COCO benchmark, our proposed approach achieves 36.6 AP$_{50}$ on novel classes, an absolute 8.2 gain over the previous best performance. For LVIS, we surpass the state-of-the-art ViLD model by 5.0 mask AP for rare categories and 3.4 overall. Code: \url{https://github.com/hanoonaR/object-centric-ovd}.
\extrafootertext{\textsuperscript{*}Equal contribution}
\end{abstract}

\section{Introduction}

Open-vocabulary detection (OVD) aims to generalize beyond the limited number of base classes labeled during the training phase. The goal is to detect novel classes defined by an unbounded (open) vocabulary at inference. Owing to the challenging nature of the OVD task, different forms of weak-supervision for novel categories are typically used, \emph{e.g.}, extra image-caption pairs to enlarge the vocabulary \cite{zareian2021open}, image-level labels on classification datasets \cite{zhou2022detecting}  and pretrained open-vocabulary classification models like CLIP \cite{radford2021learning}. The use of weak-supervision to enlarge the vocabulary is intuitive as the cost of annotating large-category detection datasets is monumental while the image-text/label pairs are readily available via large classification datasets \cite{deng2009imagenet} or internet sources \cite{radford2021learning,jia2021scaling}.

One of the major challenges with enlarging vocabulary via image-level supervision (ILS) or pretrained models learned using ILS is the inherent mis-match between region and image-level cues. For instance,  pretrained CLIP embeddings used in the existing OVD models \cite{gu2021open,zhou2022detecting} do not perform well in locating object regions \cite{zhong2021regionclip} since the CLIP model is trained with full scale images. Similarly, weak supervision on images using  caption descriptions or image-level labels does not convey the precise object-centric information. For label grounding in images, the recent literature explores expensive pretraining with auxiliary objectives \cite{zareian2021open} or use heuristics such as, the max-score or max-size boxes \cite{zhou2022detecting}.

In this paper, we set out to bridge the gap between object and image-centric representations within the OVD pipeline. To this end, we propose to utilize high-quality {class-agnostic} and class-specific object proposals via the pretrained multi-modal vision transformer (ViT) \cite{maaz2021multi}. 
The  \emph{class-agnostic} object proposals are then used to distill region-specific information in the CLIP visual embeddings, making them suitable for local objects. Furthermore, the \emph{class-specific} proposal set allows us to visually ground a larger vocabulary, thereby aiding in generalization to novel categories. Next, the final and important question is how to make visual-language (VL) mapping amenable to local object-centric information. For this purpose, we introduce a region-conditioned weight transfer process which closely ties together image and region VL mapping. In a nut-shell, the proposed approach connects the image, region and language representations to generalize better to novel open-vocabulary objects.

The major contributions of this work include:\vspace{-0.5em}
\begin{itemize}\setlength{\itemsep}{0mm}
    \item We propose \emph{region-based knowledge distillation} to adapt image-centric CLIP embeddings for local regions, thereby improving alignment between region and language embeddings. We show that the resulting well-aligned representations aid in improving the overall performance of our text driven OVD pipeline. 
    \item In order to visually ground weak image labels, our approach performs \emph{pseudo-labeling} using the high-quality object proposals from pretrained multi-modal ViTs. This helps in enlarging the class vocabulary and therefore generalizes better to new object classes. 
    \item The above contributions mainly target the visual domain. In order to preserve the benefits of object-centric alignment in the language domain, we also propose to explicitly condition the (pseudo-labeled) image-level VL mapping on the region-level VL mapping via a novel \emph{weight transfer function}. In this manner, we are the first to simultaneously integrate object-centric visual and language alignment within a single architecture for OVD. 

    \item Our extensive experiments demonstrate the improved OVD capability of the proposed approach. On COCO and LVIS benchmarks, our method achieves absolute gains of 8.2 and 5.0 AP on novel and rare classes over the current SOTA methods. Further generalizability is demonstrated by our cross-dataset evaluations performed on COCO, OpenImages and Objects365, leading to consistent improvements compared to existing methods.

\end{itemize}

\section{Related Work}
\textbf{Zero-shot Object Detection (ZSD):} This setting involves detecting novel class objects at inference, for which no visual examples are available during training. Zhu~\etal~\cite{zhu2019zero} use semantic information with visual features to get proposals for both seen and unseen classes. Bensal~\etal~\cite{bansal2018zero} show that learning a good separation between background and foreground is critical in ZSD and propose to use multiple latent classes for modeling background during training. Rahman~\etal~\cite{rahman2020improved} propose a polarity loss to solve the ambiguity between background and unseen classes. DELO~\cite{zhu2020don} focuses on generating good proposals for unseen classes by synthesizing visual features for unseen objects using a generative model. Gupta~\etal~\cite{gupta2020multi} benefits from the contemporary cues in semantic and visual space ensuring better class separation for ZSD. Other works use additional learning signals, including unlabeled images from target domain~\cite{rahman2019transductive} and raw textual descriptions from the internet~\cite{li2019zero}.
Although significant progress has been made on this topic~\cite{rahman2019transductive, li2019zero, gupta2020multi}, the inherent complexity of the task makes it challenging for the ZSD models to generalize well to unseen object classes. 

\textbf{Weakly-supervised Object Detection (WSOD):}
In this setting, only image-level labels are used to approach object detection~\cite{shen2020enabling, shen2019cyclic, wan2019c, yang2019towards, zhong2020boosting}, or are used alongside the detection dataset to enlarge the detector vocabulary~\cite{yan2017weakly, dong2021boosting, fang2021wssod}. Bilen~\etal~\cite{bilen2016weakly} proposed a weakly-supervised deep detection network (WSDNN) that uses off-the-shelf region proposals~\cite{uijlings2013selective, zitnick2014edge} and computes objectness and recognition scores for each proposal using separate subnetworks. Cap2Det~\cite{ye2019cap2det} operates in a similar setting and uses raw text captions to generate pseudo-labels to guide image-level supervision. Li~\etal~\cite{li2019weakly} uses segmentation-detection collaborative network (SDCN) for accurate detection under weakly-supervised setting using only image labels. PCL~\cite{tang2018pcl} proposes to cluster the spatially adjacent proposals and then assign image labels to each cluster. CASD~\cite{huang2020comprehensive} argues that the detectors trained only with image-level labels are prone to detect boxes around salient objects and propose feature attention along with self-distillation to address the issue. YOLO9000~\cite{redmon2017yolo9000} and DLWL~\cite{ramanathan2020dlwl} augments the detection training by assigning image-level labels to the max-score proposal. Detic~\cite{zhou2022detecting} shows that using max-size proposal is an optimal choice for assigning image-level labels as it does not rely on the predictions of the network being optimized and provides better signals for the novel classes. 
We also operate in a similar WSOD setting and use high-quality object proposals from pretrained multi-modal ViT~\cite{maaz2021multi} to enlarge detector vocabulary and generalize towards novel object categories.

\textbf{Open-vocabulary Object Detection (OVD):}
In OVD, the objective is to detect target class objects not present in the training/base class vocabulary. 
A typical solution of the problem is to replace the classifier weights with text embeddings of the target vocabulary (\eg, GloVe~\cite{pennington2014glove}, BERT~\cite{devlin2018bert}, CLIP~\cite{radford2021learning}).
OVR-RCNN~\cite{zareian2021open} uses BERT embeddings as classifier weights and proposes to use open-vocabulary captions to learn the vision-to-language mapping. It surpasses the ZSD approaches by a large margin. ViLD~\cite{gu2021open} uses pretrained CLIP~\cite{radford2021learning} to distill knowledge into a two-stage object detector~\cite{ren2015faster} and replaces the classifier weights with CLIP text embeddings obtained by ensembling multiple text prompts (\eg, \txt{a \{category\}, a photo of a \{category\}}). Gao~\etal~\cite{gao2021towards} generate pseudo bounding-box labels using pretrained VL models for training open-vocabulary detector. All these methods use carefully designed manual prompts for generating text embeddings. DetPro~\cite{du2022learning} and PromptDet~\cite{feng2022promptdet} replace these manual prompts with learnable tokens and achieve competitive results on novel/rare categories. However, in our work, we use fixed manual prompts and instead focus on improving the object-centric representations for open-vocabulary object detection.
\vspace{-0.3cm}

\section{Object-centric Open-Vocabulary Detection}
\begin{figure*}[!t]
\centering
{\includegraphics[width=0.98\textwidth]{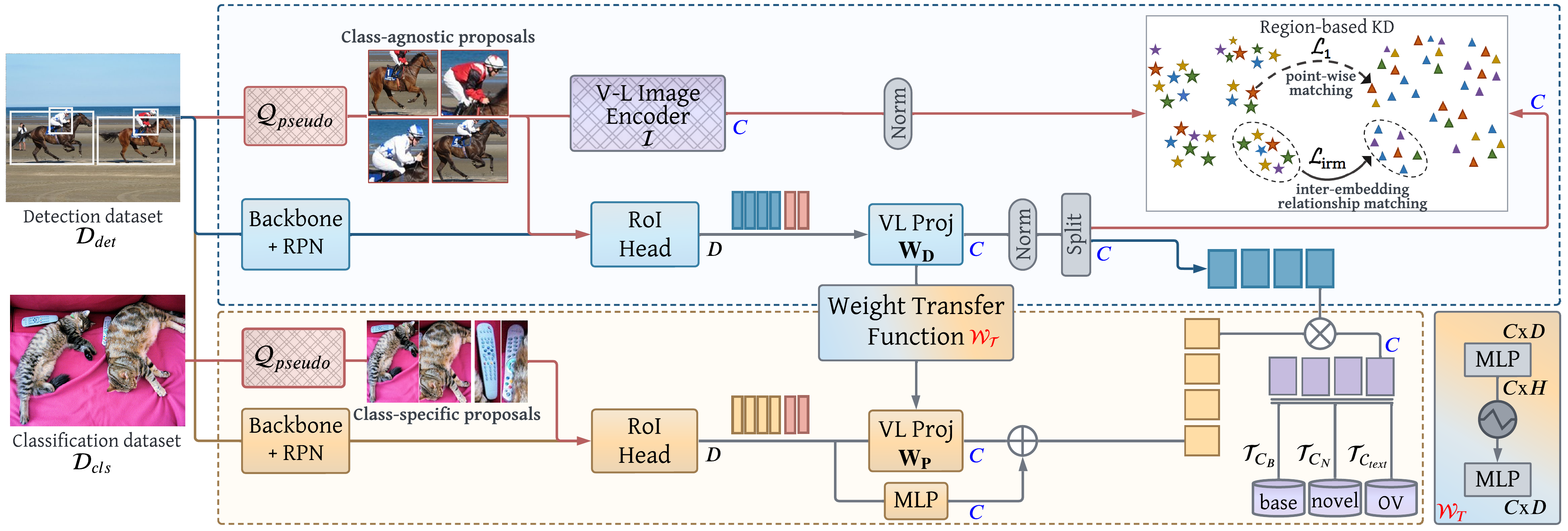}}
\caption{\small\textbf{An overview of our proposed object-centric framework for OVD.} We pair a two-stage object detector with fixed language embeddings from a pretrained  \colorbox{lavender}{visual-language (VL) model, CLIP~\cite{radford2021learning}}.  Our proposed \colorbox{lightred}{pseudo-labeling strategy $\mathcal{Q}_{\text{pseudo}}$} uses pretrained multi-modal ViTs to obtain high-quality class-agnostic and class-specific proposals. The overall pipeline follows a stage-wise learning strategy. \emph{First,} we introduce \colorbox{lightblue}{region-based knowledge distillation (RKD)} to adapt image-centric CLIP embeddings for local regions. Using the pretrained VL image encoder as a teacher model, we train the detector to induce point-wise and inter-embedding relationship alignment with our region embeddings using class-agnostic proposals from $\mathcal{Q}_{\text{pseudo}}$. \emph{Next,} we utilize a \colorbox{lightyellow}{weakly-supervised} learning framework by combining instance-level labels from detection dataset and image-level labels from classification dataset which are visually grounded using $\mathcal{Q}_{\text{pseudo}}$. This weak-supervision helps in enlarging the class vocabulary and generalizes the detector to novel classes. To preserve the benefits of object-centric alignment in the language domain learned via RKD, we explicitly condition the image-level VL mapping $W_P$, on the learned region-level VL mapping $W_D$ via a novel weight transfer function.}
\label{ovd_block_diag}
\end{figure*}

Here, we first present a brief overview of the proposed open-vocabulary detection (OVD) framework. As discussed earlier, existing OVD methods use different forms of weak supervision that employ image-centric representations, making them less suited for the end detection task. Our proposed method aims to bridge the gap between image and object-centric visual-language (VL) representations. We summarize the architectural overview of our method in Fig.~\ref{ovd_block_diag}. The proposed design has three main elements. 1) Our \emph{region-based knowledge distillation} (refer Sec.~\ref{RKD}) adapts image-centric language representations to be object-centric. A VL mapping learns to align the local region representations of the detector to the language representations by distilling the detector's region representations with region representations from a VL model (CLIP). 2) Given weak image-level supervision, we use \emph{pseudo-labeling} from pretrained multi-modal ViTs (refer Sec. \ref{ILS}) to improve generalization of the detector to novel classes. 3) For an efficient combination of the above two proposed components, we condition the VL mapping learned during the weak supervision on the VL mapping learned with region-based distillation via a novel \emph{weight transfer function} (refer Sec.~\ref{WT}). Specifically, we follow a stage-wise learning strategy to first align the region and language embeddings using RKD, and use this distilled VL mapping for object-centric visual and language alignment in the subsequent stage.

\subsection{Detection Pipeline: Preliminaries}\label{sec1}
In the open-vocabulary detection problem, we have access to an object detection dataset where the training set, $\mathcal{D}_{\text{det}}$, comprises samples from the set of base object categories, $\mathcal{C}_{\text{B}}$. The images of $\mathcal{D}_{\text{det}}$ are exhaustively annotated with bounding-box labels and corresponding class labels $y_r \in {\mathcal{C}_{\text{B}}}$, for the different objects in the image. Given an image $I \in \mathbb{R}^{H \times W \times 3}$, we design an open-vocabulary object detector to solve two subsequent problems: (1) effectively localize all objects in the image, (2) classify the detected region into one of the class label of $\mathcal{C}_{\text{test}}$, which is provided by the user at test time. The categories during test time also include novel categories $\mathcal{C}_{\text{N}}$ beyond the closed set of base categories seen during the training phase, \ie, $\mathcal{C}_{\text{test}} = \mathcal{C}_{\text{B}} \cup \mathcal{C}_{\text{N}}$. 

We convert a generic two-stage object detector \cite{ren2015faster} to an open-vocabulary detector by replacing the learnable classifier head with fixed language embeddings, $\mathcal{T}$ corresponding to the category names of $\mathcal{C}_{\text{test}}$, that are obtained using a large-scale pretrained VL model.
Following~\cite{gu2021open}, we use the \emph{text embeddings} from CLIP text encoder \cite{radford2021learning} for classification, where only the embeddings of $\mathcal{C}_{\text{B}}$ categories, $\mathcal{T}_{\mathcal{C}_{\text{B}}}$ are used during training. Specifically, we generate the text embeddings offline, by processing the prompts corresponding to each category with a template of ‘\txt{a photo of \{category\}}’ through the CLIP text encoder. The RoI \cite{ren2015faster} head computes pooled feature representations $\phi(r)$ of the proposals $r$ generated by the region proposal network (RPN). These feature embeddings are projected to a common feature space shared by the text embedding $\mathcal{T}$ using a linear layer $f(\cdot)$, which we represent as \emph{region embeddings}, $\mathcal{R} = f(\phi(r)) \in \mathbb{R}^{D}$. For classification, we compute the cosine similarity between the region embeddings and text embeddings to find the matching pairs. During training, the regions that do not match with any of the ground-truths are assigned to the background category represented by a fixed all zero embedding. We compute the cosine similarity by comparing each region to each base class,
$\mathcal{V} = {sim}(r, b) = \cos \big(\mathcal{R}(r),  \mathcal{T}_b \big) \ \forall \  b \in \mathcal{C}_{\text{B}}$.  The classification loss is a softmax cross-entropy (CE) where the logits are the cosine similarity scores, 
\begin{align*}
    \mathcal{L}_{cls} = \frac{1}{N} \sum_{r} \mathcal{L}_{CE} \left({\txt{softmax}}\Big(\frac{\mathcal{V}}{\tau} \Big), y_r\right ), \  \  y_r \in \mathcal{C}_{\text{B}}.
\end{align*}
where $\tau$ is the temperature, $N$ is the total number of proposals per image, and $r$ represents a single proposal with the ground-truth label $y_r$. 

\subsection{Region-based Knowledge Distillation}\label{RKD}
In the OVD setting, we assume that $f(\cdot)$ learns a VL mapping and aligns the output region embeddings of the detector with the corresponding CLIP text embeddings.
However, the performance on novel categories is not comparable to what CLIP encoded embeddings would provide~(refer Appendix~\ref{appendix_zero_shot} for details). We hypothesize that this performance gap is mainly due to two reasons, \romannum{1}) the data that has been used for training CLIP model 
consist of scene-centric images, making it less suitable for region classification, \emph{e.g.,} in our case where object-centric tightly bounded proposals are used, \romannum{2}) the zero-shot generalization ability of the pair-wise trained CLIP image and text embeddings cannot be fully utilized due to the mismatch between regions representations from CLIP image encoder and our detector. Based on these insights, we propose a \textbf{\textit{region-based knowledge distillation} (RKD)}.  

The proposed RKD uses distillation in the detection pipeline by distilling region embeddings from high-quality class-agnostic proposals ($\Tilde{r}$) obtained from a pretrained multi-modal ViT (MViT)~\cite{maaz2021multi}. Note that we obtain both class-agnostic (used in RKD) and class-specific (refer Sec.~\ref{ILS}) object proposals using this pseudo-labeling process, which we refer to as $\mathcal{Q}_{\text{pseudo}}$.  This is possible via using intuitive text queries to interact with the MViT model that can locate generic objects and provides the corresponding set of candidate proposals. The queries can be generic or targeted, based on the task, \eg, ‘\txt{all objects}’ to generate class-agnostic proposals, or ‘\txt{every dog}’ for a specific class.

For RKD, we compute class agnostic proposals on $\mathcal{D}_{\text{det}}$ using simple text query, ‘\txt{all objects}’ and select top-K  proposals (Fig.~\ref{fig:rpn_vs_mvit}b). CLIP embeddings $\mathcal{I}(\Tilde{r})$ are then computed offline using the CLIP image encoder $\mathcal{I}(\cdot)$. With the detector region embeddings and the corresponding CLIP region representations, we propose to use two types of distillation losses to improve the alignment.

\textbf{(1)} \textbf{\emph{Point-wise embedding matching loss:}} The $\mathcal{L}_1$ loss matches the individual region embeddings $\mathcal{\Tilde{R}} = f(\phi(\Tilde{r}))$ with the CLIP region representations $\mathcal{I}(\Tilde{r})$, 
\begin{gather} 
\mathcal{L}_1 = \frac{1}{K} \sum_{\Tilde{r}}  \parallel \mathcal{\Tilde{R}}  - \mathcal{I}(\Tilde{r}) \parallel_{1}.
\label{l1_loss}
\end{gather}
Using this criteria, our visual encoder, along with the VL projection layer $f(\cdot)$, approximates the CLIP image encoder and consequently aligns our region embeddings with the CLIP text embeddings.

\begin{figure*}[!t]
\centering
{\includegraphics[width=1.0\textwidth]{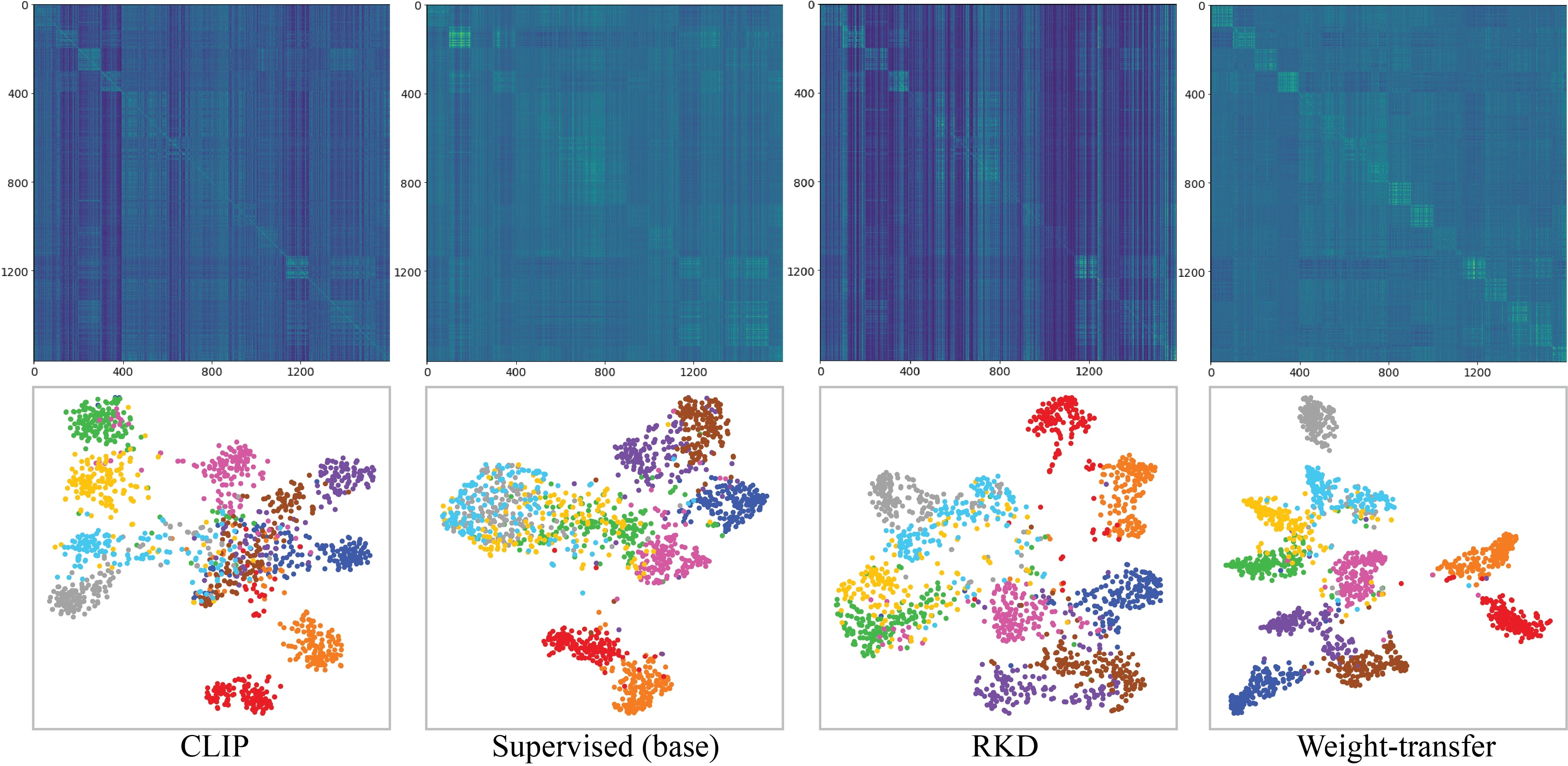}}
\caption{\small \emph{Top-row}: Similarity matrices computed on the CLIP (${S_I}$) and detector (${S_R}$) region embeddings for COCO novel classes. A subset of 100 randomly selected samples per category form a batch represented by a column are grouped together. Our region-based distillation enforces the similarity patterns in the RKD model to be closer to the teacher model, CLIP, indicated by the bright colors along diagonals. \emph{Bottom-row}: t-SNE plots of CLIP and detector region embeddings on novel COCO categories. The CLIP aligned RKD and weight transfer detector embeddings shows improved separability among novel class features as compared to the supervised detector region embeddings \emph{(figure best viewed in-zoom).}}
\label{fig:spkd_tsne}
\end{figure*}

\textbf{(2)} \textbf{\emph{Inter-embedding relationship matching loss (IRM):}} It is a knowledge distillation based loss  $\mathcal{L}_{irm}$ that instills inter-embedding relationships within our region representations to be consistent to the CLIP region representations  \cite{Tung_2019_ICCV}. Instilling such inter-embedding relations would be beneficial as we know that the teacher model $\mathcal{I}(\cdot)$, and the student model (our detector), are different in nature with respect to their training methods (Fig.~\ref{fig:spkd_tsne}). The IRM loss is defined on pairwise similarity matrices of the two different sets of embeddings. Specifically, with the top-K proposals computed from $\mathcal{Q}_{\text{pseudo}}$, we compose $K \times K$ similarity matrices for $\mathcal{I}(\Tilde{r})$ and $\Tilde{\mathcal{R}}$ denoted by ${S_I}$ and ${S_R}$ respectively. Notably, these matrices are normalized by L2 norm applied row-wise. The IRM loss is a Frobenius norm $\parallel \cdot \parallel_F$, over the mean element-wise squared difference between ${S_{\mathcal{I}}}$ and ${S_R}$, 
\begin{gather} 
{S_R} = \frac{\Tilde{\mathcal{R}} \cdot \Tilde{\mathcal{R}}^{T}}{\parallel \Tilde{\mathcal{R}} \cdot \Tilde{\mathcal{R}}^{T} \parallel_2}, \  \  \   {S_{\mathcal{I}}}= \frac{\mathcal{I}(\Tilde{r}) \cdot \mathcal{I}(\Tilde{r})^{T}}{\parallel {\mathcal{I}(\Tilde{r}) \cdot \mathcal{I}(\Tilde{r})^{T}} \parallel_2}, \nonumber \\ 
\mathcal{L}_{{irm}} = \frac{1}{K^2} \parallel {S_R} - {S_{\mathcal{I}}} \parallel_F^2.
\label{l_IRM}
\end{gather}
We weight the $\mathcal{L}_{1}$ and $\mathcal{L}_{irm}$ losses by factors $\beta_1$ and $\beta_2$, respectively. Together with the standard two-stage detector losses; RPN loss ($\mathcal{L}_{rpn}$), regression loss ($\mathcal{L}_{reg}$) and classification loss ($\mathcal{L}_{cls}$)~\cite{ren2015faster, he2017mask}; the overall training objective with RKD can be expressed as, 
\begin{equation}
    \mathcal{L}_{RKD} = \mathcal{L}_{rpn} + \mathcal{L}_{reg} + \mathcal{L}_{cls} + \beta_1 \ \mathcal{L}_{1} + \beta_2 \ \mathcal{L}_{irm}.
\label{RKD_loss}
\end{equation}

\vspace{-1em}
\subsection{Image-level Supervision with Pseudo Box Labels}\label{ILS}
\begin{figure*}[!t]
\centering
{\includegraphics[width=1.0\textwidth]{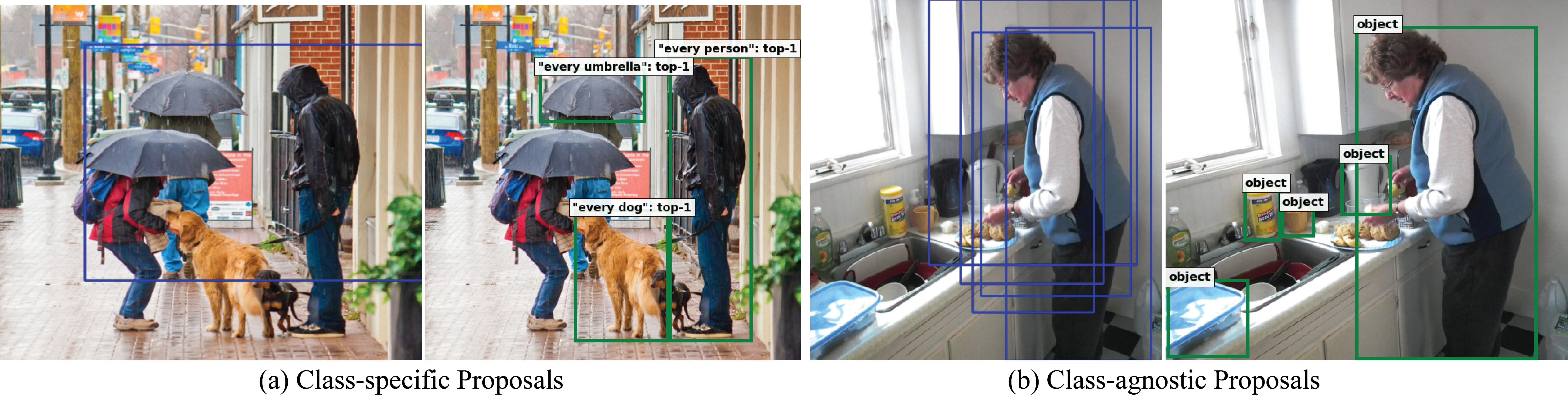}}
\caption{\small \textbf{\color{blue}{(a)}} \textbf{Class-specific Proposals:} A visual comparison of heuristic methods (\emph{left}) used for visual grounding in image-level supervision \cite{zhou2022detecting} with our proposed method (\emph{right}). Using heuristic based approaches like selecting maximum sized box from the RPN can ignore local objects in the scene. In our method, we design class-specific text queries with known class labels for pseudo-labeling potential objects. \textbf{\color{blue}{(b)}} \textbf{Class-agnostic Proposals:} In region-based knowledge distillation (RKD), we induce better region-level alignment with fewer high-quality proposals from a generalized class-agnostic proposal generator \cite{maaz2021multi}. We compare top-K RPN proposals (\emph{left}) with top-K multi-modal ViTs proposals used in a class-agnostic manner (\emph{right}).}
\label{fig:rpn_vs_mvit}
\end{figure*}

In the open-vocabulary setting, a fundamental challenge is to generalize the detector to novel classes. However, due to the daunting task of densely locating all objects in natural scenes, the existing detection datasets are of relatively smaller magnitude compared to the classification datasets, which are easier to annotate. To this end, Zhou \emph{et al.} \cite{zhou2022detecting} proposed to take advantage of a large-scale image classification dataset during training to expand the detector's vocabulary. However, an important question is how to effectively associate the region proposals of novel objects with the corresponding labels.  We note that the existing approach uses heuristics such as selecting the whole image as a single box, or just the maximum sized box from the RPN, which can ignore potential objects (Fig.~\ref{fig:rpn_vs_mvit}a).

We propose a weakly-supervised method to generalize the detector to novel categories by using pseudo-box labels from pretrained MViT~\cite{maaz2021multi}. We follow \cite{zhou2022detecting} to train the detector with a combination of detection and classification dataset. A batch of data is prepared by combining data from the detection dataset $\mathcal{D}_{\text{det}}$ that are exhaustively annotated with bounding-box and class labels, with data from a classification dataset $\mathcal{D}_{\text{cls}}$ that only contains image-level labels. With $\mathcal{Q}_{\text{pseudo}}$, we obtain the pseudo-box labels on this classification dataset, which we use for \textbf{\textit{image-level supervision} (ILS)}. 
Specifically, consider a sample image $I \in \mathcal{D}_{\text{cls}}$, which has a total of $N$ ground-truth class labels, we generate object proposals offline with the use of MViT corresponding to these weak labels. Specifically, we construct $N$ class-specific text queries $\{t_n\}_{n=1}^N$ with template ‘\txt{every \{category\}}’, and obtain $K$ proposals $\{\Tilde{r}_k\}_{k=1}^{K}$ and corresponding confidence scores $\{\Tilde{s}_k\}_{k=1}^{K} $ for each query,
\begin{align*}
    [(\Tilde{r_1}, \Tilde{s_1}), (\Tilde{r_2}, \Tilde{s_2}),  \cdots (\Tilde{r_K}, \Tilde{s_K})] =  \mathcal{Q}_{\text{pseudo}}(I, t_n);  \ \ I\in \mathcal{D}_{\text{cls}}, n \in N .\nonumber
\end{align*}
We select the top-1 proposal with the highest confidence score, as the pseudo-box label for a particular category. This gives us $N$ high-quality pseudo-box labels for each image, corresponding to its $N$ image-level category labels~(Fig.~\ref{fig:rpn_vs_mvit}a). We compute the region embeddings $\mathcal{\Tilde{R}}$ for proposals $\Tilde{r}$ as,
\begin{align*}
    \mathcal{\Tilde{R}}_n = f(\phi({\Tilde{r}_{\hat{k}}})), \ \ \hat{k}= \text{argmax}_k (\Tilde{s_k}). \nonumber
\end{align*}
In the case of $\mathcal{D}_{\text{det}}$, the training follows the standard two-stage RCNN training recipe. However, for $\mathcal{D}_{\text{cls}}$, only the classification loss is updated. We call this \textit{pseudo-max score}, $\mathcal{L}_{pms}$ loss.
\begin{gather} 
\mathcal{L}_{pms} = \frac{1}{N} \sum_{n} BCE(\mathcal{V}, y_{\Tilde{r}}), \text{ where } \mathcal{V} = \cos \big(\mathcal{\Tilde{R}}_n,  \mathcal{T} \big).
\label{l_pms}
\end{gather}
We weight $\mathcal{L}_{pms}$ by a factor $\alpha$ and the overall training objective with our ILS can be expressed as,
\begin{equation}
    \mathcal{L}_{ILS} =  \left\{ \begin{array}{lcr}
\mathcal{L}_{rpn} + \mathcal{L}_{reg} + \mathcal{L}_{cls}, & \mbox{if} & I \in \mathcal{D}_{\text{det}} \\
\alpha \ \mathcal{L}_{pms}, & \mbox{if} & I \in \mathcal{D}_{\text{cls}}.\\
\end{array}\right.
\label{ILS_loss}
\end{equation}

\vspace{-1em}
\subsection{Weight Transfer Function}\label{WT}
To combine the alignment from region-based distillation (Sec.~\ref{RKD}) with the benefits from weak supervision with pseudo-box labels (Sec.~\ref{ILS}), a naive approach would be to train the detector with a combination of losses: $\mathcal{L}_{1}$ (\ref{l1_loss}), $\mathcal{L}_{irm}$ (\ref{l_IRM}) and $\mathcal{L}_{pms}$ (\ref{l_pms}). However, we demonstrate that a simple combination of the two approaches does not lead to complimentary benefits, instead they compete with each other (Table~\ref{tab:our_approach_summary}). The additional supervision from pseudo-labels improves the generalization of the detector, while the region-based distillation works towards object-centric alignment in the language domain, thereby improving the overall performance of the detector. We aim to incorporate the benefits from the two approaches and preserve the object-centric alignment in the language domain. To this end, we use a weight transfer mechanism \cite{hu2018learning} from VL projection used in region-based distillation to the weak supervision by learning a \textbf{\textit{weight transfer function}}, $\mathcal{W_T(\cdot)}$. In other words, the VL projection function $f(\cdot)$ used during the weak image-level supervision is explicitly conditioned on the mapping function used for alignment in the distillation process. This way, both the transformations are tied together to reinforce mutual representation capability and avoid any conflict in the learned function mapping. Let the weights of the projection layer in RKD and weak image-level supervision be represented as $W_D$ and $W_P$ respectively. The weight transfer operation is given by,
\begin{align*}
    W_P =  \mathcal{W_T}(W_D) = \Big( W_{\theta_2} \ \rho (W_{\theta_1} \ W_{D}) \Big) &; \ \ \ \ \ \ \ \ \ \mathcal{W_T}\colon \ W_D \rightarrow W_P.
\end{align*}
Here, $W_D$ is kept frozen and we design $\mathcal{W_T}$ as a 2-layer MLP, $W_{\theta_1}$ followed by $W_{\theta_2}$ a with LeakyReLU ($\rho$) activation with a negative slope of 0.1. Further, we use a skip connection across $W_P$ by projecting the original representations using a separate 2-layer MLP (Fig.~\ref{ovd_block_diag}).
The total loss here is a combination of $\mathcal{L}_{RKD}$~(Eq.~\ref{RKD_loss}) and $\mathcal{L}_{ILS}$~(Eq.~\ref{ILS_loss}) loss, given by,
\begin{gather*}
    \mathcal{L} = \mathcal{L}_{rpn} + \mathcal{L}_{reg} + \mathcal{L}_{cls} + \beta_1 \ \mathcal{L}_{1} + \beta_2 \ \mathcal{L}_{irm} + \alpha \ \mathcal{L}_{pms}.
\end{gather*}

\vspace{-2em}
\section{Experiments}\label{experiments_section}

\subsection{Datasets}\label{checklist_dataset_details}

\begin{wraptable}{r}{8cm}
\vspace{-0.15in}
\setlength{\tabcolsep}{4pt}
\resizebox{1.0\linewidth}{!}{
\begin{tabular}{lcccccc}
\toprule
\multirow{1}*{Dataset} & \multirow{1}*{Dataset Type} & Task & \multirow{1}*{\# images} \\
\midrule
COCO & Detection & OVD  & 118K\\
LVIS v1.0 & Detection & OVD & 100K\\
ImageNet-21K$^*$ & Classification & ILS in LVIS & 1.4M\\
COCO-Captions & Image-captioning & ILS in COCO & 118K\\
\midrule
\multirow{2}*{LMDet} & Flickr30, GQA \&  & \multirow{1}* {MViT} & \multirow{2}* {1.1M}\\
                     & \multirow{1}* Visual Genome & Pretraining &\\
\midrule
\multirow{3}*{$\ddag$ LMDet} & LMDet & \multirow{1}* {MViT} & \multirow{3}* {0.8M}\\
                     & \multirow{1}* (excluding any overlap & Pretraining &\\
                     & \multirow{1}* with novel categories) &  &\\
\bottomrule
\end{tabular}
}
\caption{Summary of the datasets used in our experiments.}
\label{datasets}
\vspace{-0.2in}
\end{wraptable}

We conduct our experiments on COCO \cite{coco} and LVIS v1.0 \cite{gupta2019lvis} under OVD setting. For evaluation, we use the generalized ZSD setting where the classifier contains both base and novel categories. Table~\ref{datasets} summarizes all the datasets used in our work. Following~\cite{zhou2022detecting, zareian2021open}, we use a subset of ImageNet-21K having 997 overlapping LVIS categories and COCO captions dataset for ILS in LVIS and COCO experiments respectively (refer Appendix.~\ref{appendix:impl_details} for more details). For the pseudo-labeling process $\mathcal{Q}_{\text{pseudo}}$, we use the MViT pretrained on a Large-scale Modulated Detection (LMDet) dataset \cite{maaz2021multi}. We ensure that MViT pretraining dataset has no overlap with any of the evaluation datasets in our work. Additionally, in all our experiments we use a pretrained MViT that we train using the author's provided code on filtered LMDet ($\ddag$LMDet) dataset by entirely restricting any exposure to the novel/rare classes in evaluation.

\textbf{COCO OVD:}
We use COCO-2017 dataset for training and validation. We follow the ZS splits proposed in \cite{bansal2018zero}, in which 48 categories are selected as base and 17 are selected as novel classes. 

\textbf{LVIS OVD:}
LVIS contains 1203 categories which are further split into frequent, common and rare categories. Inline with~\cite{gu2021open, zhou2022detecting}, we combine the frequent and common categories to form base classes and keep all rare classes as novel, resulting in 866 base and 337 rare classes. 

\textbf{Cross-transfer Datasets:}
To validate the adaptability of our method, we evaluate and compare results of our LVIS trained model on OpenImages\cite{kuznetsova2020open} and Objects365~\cite{shao2019objects365} and COCO~\cite{coco} datasets.

\subsection{Implementation details}\label{checklist_implementation_details}
We conduct COCO experiments using Faster R-CNN~\cite{ren2015faster} with ResNet-50 backbone. We train the supervised-base model on 48 base classes ($\mathcal{C}_{\text{B}}$) for 1x schedule ($\sim$12 COCO epochs)
and report box AP$_{50}$. For RKD, we finetune this model for another 1x schedule using box labels from $\mathcal{C}_{\text{B}}$ and class-agnostic proposals from the pretrained MViT~\cite{maaz2021multi}. This model is further finetuned for 1x schedule with ILS and the associated weight transfer function using class labels from COCO captions and corresponding class-specific proposals from MViT. This sums to an overall 3x training schedule.

For LVIS experiments, we use Mask R-CNN~\cite{he2017mask} with federated loss~\cite{zhou2021probabilistic} and sigmoid cross-entropy, and report mask AP. For RKD and weight transfer, we use the same training schedules as of COCO and report the average over three runs. For comparison with Detic~\cite{zhou2022detecting}, we apply our proposed method on their strong CenterNetV2~\cite{zhou2021probabilistic} baseline under the same settings. It uses ImangeNet21K pretrained backbone with 4x schedule using large scale jittering (LSJ)~\cite{lsj} augmentations. All of our models are trained using 8 A100 GPUs with an approximate training time of 9 and 6 hours for 1x schedule of COCO and LVIS respectively.

In our experiments, we use SGD optimizer with a weight decay of $1e^{-4}$ and a momentum of 0.9. We train for 1x schedule with batch size of 16 and an initial learning rate of 0.02 which drops by a factor of 10 at the 8$^{th}$ and 11$^{th}$ epoch. We set temperature $\tau$ to 50. Our longer schedules experiments use 100-1280 LSJ~\cite{lsj}. We use $\alpha$ of 0.1 to weight $\mathcal{L}_{pms}$. For computing CLIP embeddings we use the CLIP model ViT-B/32~\cite{radford2021learning}, with input size of 224$\times$224. We use the query ‘\txt{a photo of a \{category\}}’ for to compute the text embeddings for the classifier. For distillation, we use top 5 proposals from the pretrained MViT~\cite{maaz2021multi} evaluated with generic query, \lq\txt{all objects}\rq, generating class-agnostic proposals. We refer to Appendix~\ref{appendix:mvit} for additional details on the approach we use to generate class-agnostic and class-specific proposals from MViT. In COCO experiments, we set weights $\beta_1$ and $\beta_2$ to 0.15. In LVIS, we set $\beta_1$ to 0.15 and $\beta_2$ to 0.25. We choose these values using a randomized hyper-parameter search on the corresponding held-out datasets. The 2-layer MLP in our weight transfer function has a hidden dim of 512, and a hidden dim of 1024 is used in the MLP skip connection across $W_P$ in Fig.~\ref{ovd_block_diag} (refer to Appendix \ref{appendix:ablations} for more details).

\subsection{Our Approach: Main results}
Table~\ref{tab:our_approach_summary} shows the contribution of individual components in our proposed approach. Building on top of the supervised-base model, our \textit{region-based knowledge distillation} (RKD) shows an absolute gain of 19.5 and 1.5 AP for COCO novel and base classes respectively, indicating the adaptability of image-centric CLIP embeddings for local regions. 
With \textit{pseudo-box labeled weak image-level supervision} (PIS), novel class AP improves by 28.7, demonstrating generalization to novel classes and thus enlarging the detector's vocabulary. 
Naively combining the two approaches shows improvement, but struggles to maintain the gains from the individual components. In contrast, our \textit{weight transfer} method suitably combines the complimentary benefits of both components~(Fig.~\ref{fig:spkd_tsne}), achieving 36.6 AP on novel classes while maintaining performance on base classes.

\begin{SCtable}[\sidecaptionrelwidth][h!]
\hspace{-0.6em}
\centering
\resizebox{0.68\linewidth}{!}{
\begin{tabular}{lccccc}
\toprule
\multirow{1}*{Method} & \multirow{1}*{AP$_{\txt{novel}}$} & \multirow{1}*{AP$_{\txt{base}}$} & \multirow{1}*{AP} \\
\midrule
1: Supervised (Base) & 1.7	& 53.2	& 39.6 \\
\midrule
2: Base + Region based ditillation (RKD) & 21.2 & \textbf{54.7} & 45.9 \\
3: Base + ILS with pseudo-box (PIS) & 30.4 & 52.6 & 46.8\\
4: RKD + PIS & 31.5 & 52.8 & 47.2 \\
5: RKD + PIS + Weight-transfer (Ours) & \textbf{36.6} & 54.0 & \textbf{49.4} \\ \bottomrule
\end{tabular}
}
\hspace{-1.3em}
\caption{Effect of individual components in our method. Our weight transfer method provides complimentary gains from RKD and ILS, achieving superior results as compared to naively adding both components.}
\label{tab:our_approach_summary}
\end{SCtable}

\begin{table}[!b]
\vspace{-0.1cm}
\begin{center}
\small
\setlength\tabcolsep{4pt}
\begin{tabular}{lcccccc}
\toprule
\multirow{1}*{Method} & Supervision & \multirow{1}*{AP$_{\txt{base}}$} & \multirow{1}*{AP$_{\txt{novel}}$} & \multirow{1}*{AP} \\
\midrule
WSDDN$\S$~\cite{bilen2016weakly}                 & \multirow{2}* {image-level labels for $\mathcal{C}_{\text{B}} \cup \mathcal{C}_{\text{N}}$}   & 19.6 & 19.7 & 19.6 \\
Cap2Det$\S$~\cite{ye2019cap2det}                 &                                                         & 20.1 & 20.3 & 20.1  \\ \arrayrulecolor{grey}\hline

\multirow{2}*{OVR-CNN~\cite{zareian2021open}}& \multirow{1}*{pretraining with captions $\mathcal{C}_{\text{B}} \cup \mathcal{C}_{\text{N}}$ }&\multirow{2}*{46.0}& \multirow{2}*{22.8}& \multirow{2}*{39.9}\\
                                             & \multirow{1}*{box-level labels in $\mathcal{C}_{\text{B}}$}                & & &\\ \arrayrulecolor{grey}\hline
                                             
\multirow{2}*{ViLD\dag~\cite{gu2021open}}&\multirow{1}*{internet sourced image-text pairs}&\multirow{2}*{\textcolor{darkgrey}{\textbf{59.5}}}& \multirow{2}*{\textcolor{darkgrey}{27.6}}& \multirow{2}*{\textcolor{darkgrey}{51.3}} \\
                                             & \multirow{1}*{box-level labels in $\mathcal{C}_{\text{B}}$}                & & &\\ \arrayrulecolor{grey}\hline

\multirow{3}*{RegionCLIP~\cite{zhong2021regionclip}}&\multirow{1}*{internet sourced image-text pairs}&\multirow{3}*{{54.8}}& \multirow{3}*{{26.8}}& \multirow{3}*{{47.5}} \\
                                             & \multirow{1}*{pretraining with pseudo box-level labels }                & & &\\ \arrayrulecolor{grey}
                                             & \multirow{1}*{box-level labels in $\mathcal{C}_{\text{B}}$}                & & &\\ \arrayrulecolor{grey}
                                             \hline                                             
                                             
\multirow{2}*{Detic~\cite{zhou2022detecting}}& \multirow{1}*{internet sourced image-text pairs}  &\multirow{2}*{47.1}&\multirow{2}*{27.8}&\multirow{2}*{45.0}\\
\multirow{2}*{Detic\ddag} & \multirow{1}*{image-level labels for $\mathcal{C}_{\text{B}} \cup \mathcal{C}_{\text{N}}$} & \multirow{2}*{53.8} & \multirow{2}*{28.4}& \multirow{2}*{47.2}\\
                                              & \multirow{1}*{box-level labels in $\mathcal{C}_{\text{B}}$}                & & &\\ \arrayrulecolor{black} \hline

\noalign{\smallskip}
\multirow{2}*{Ours} &  \multirow{1}*{internet sourced image-text pairs} &\multirow{2}*{\textbf{54.0}} &\multirow{2}*{\textbf{36.6}} & \multirow{2}*{\textbf{49.4}} \\
\multirow{2}*{Ours \dag} & \multirow{1}*{image-level labels for $\mathcal{C}_{\text{B}} \cup \mathcal{C}_{\text{N}}$} & \multirow{2}*{\textcolor{darkgrey}{56.6}} & \multirow{2}*{\textcolor{darkgrey}{\textbf{36.9}}} & \multirow{2}*{\textcolor{darkgrey}{\textbf{51.5}}} \\
                                              & \multirow{1}*{pseudo-box labels in $\mathcal{C}_{\text{N}}$, box-level labels in $\mathcal{C}_{\text{B}}$}                & & &\\ \bottomrule
\end{tabular}
\vspace{0.2cm}

\caption{\textbf{OVD results on COCO.} Here $\mathcal{C}_{\text{B}}$ and $\mathcal{C}_{\text{N}}$ represents the base and novel classes respectively. $\S$The results quoted from~\cite{zareian2021open}. \dag ViLD and our methods are trained for longer 8x schedule (shown in gray). \ddag We train detic for another 1x for a fair comparison with our method.  For ViLD, we use their unified model that trains ViLD-text and ViLD-Image together. For Detic, we report their best model.}
\label{tab:coco_comparison}
\end{center}
\vspace{-0.2cm}
\end{table}

\textbf{Open-vocabulary Detection - COCO:} We compare our OVD results with previously established methods in Table~\ref{tab:coco_comparison}. OVR-CNN learns a vision-to-language mapping with expensive pretraining. Detic uses ILS to improve detection on novel classes. We use a novel weight transfer function to perform object-centric VL alignment and achieve 54.0 AP on the base classes, surpassing OVR-CNN and Detic by 8.0 AP and 0.2 AP respectively. On novel classes our method achieves 36.6 AP, the highest novel AP achieved over all methods. In comparison with ViLD, which trains for 8x schedule (${\sim}$ 96 epochs), our method with the same schedule provides 56.6 base AP, lagging by 2.9. 

On novel classes, we achieve 36.9 AP surpassing ViLD by a gain of 9.3. In contrast to ViLD design, our weight transfer function allows both RKD and ILS to provide complimentary gains without any negative competition among the two methods \cite{gu2021open}.

\textbf{Open-vocabulary Detection - LVIS:}
Table~\ref{tab:lvis_compariosn} (left) compares our results with ViLD~\cite{gu2021open} on LVIS benchmark. With 3x training schedule ($\sim$ 36 epochs) we perform reasonably well compared to ViLD 32x schedule ($\sim$ 384 epochs), already surpassing the rare AP by 1.0 while having slightly lower performance on frequent classes. Extending our model to 8x schedule fills the gap, surpassing ViLD by 0.8 in frequent and 5.0 AP in rare classes respectively. 
In Table~\ref{tab:lvis_compariosn} (right), we compare our method with Detic by using their strong LVIS baseline that uses CenterNetV2 network. Following similar settings, we finetune their box-supervised model using our weight transfer method and show improvements.

\begin{table*}[h]
\small
\setlength\tabcolsep{4.5pt}
\begin{minipage}[t]{.46\linewidth}
\centering
\begin{tabular}{lccccc}
\toprule
Method & \multirow{1}*{Epochs} & AP$_{r}$ & AP$_{c}$ & AP$_{f}$ & AP \\
\noalign{\smallskip}
\midrule
\noalign{\smallskip}
ViLD ~\cite{gu2021open}& 384 & 16.1 & 20.0 & 28.3 & 22.5 \\ \midrule
Ours & 36 & 17.1 & 21.4 & 26.7 & 22.8\\ 
Ours & 96 & \textbf{21.1} & \textbf{25.0} & \textbf{29.1} & \textbf{25.9} \\ \bottomrule
\end{tabular}
\end{minipage}
\hfill
\begin{minipage}[t]{.485\linewidth}
\centering
\setlength\tabcolsep{4.2pt}
\begin{tabular}{lcccc}
\toprule
Method & AP$_{r}$ & AP$_{c}$ & AP$_{f}$ & AP \\
\noalign{\smallskip}
\midrule
\noalign{\smallskip}
Box-supervised~\cite{zhou2022detecting} & 16.3 & 31.0 & 35.4 & 30.0 \\
Detic (Image + Captions)& 24.6 & 32.5 & 35.6 & 32.4 \\ \midrule
Ours & \textbf{25.2} & \textbf{33.4} & \textbf{35.8} & \textbf{32.9}\\ \bottomrule
\end{tabular}
\end{minipage}
\caption{\textbf{OVD results on LVIS.} (\emph{Left}): Comparison with prior work ViLD, using their unified model (ViLD-text + ViLD-Image), show improvement across novel and base categories. (\emph{Right}): We show the comparison with Detic, by building on their strong LVIS baseline using CenterNetV2 detector~\cite{zhou2022detecting}}
\label{tab:lvis_compariosn}
\vspace{-0.2cm}
\end{table*}

\begin{wraptable}{r}{5.6cm}
\vspace{-0.1in}
\setlength{\tabcolsep}{3pt}
\footnotesize
{
\begin{tabular}{lccccc}
\toprule
Method & \multirow{1}*{Epochs} & AP$_{r}$ & AP$_{c}$ & AP$_{f}$ & AP \\
\noalign{\smallskip}
\midrule
\noalign{\smallskip}
ViLD ~\cite{gu2021open}& 384 & \textbf{16.1} & 20.0 & \textbf{28.3} & \textbf{22.5} \\ \midrule
Ours & 36 & 16.0 & \textbf{20.2} & 26.3 & {21.8}\\ 
\bottomrule
\end{tabular}
}
\caption{Performance on LVIS benchmark using a strict OVD setting.}
\label{tab:strict_ovd}
\vspace{-0.2in}
\end{wraptable}
\textbf{Strict Open-vocabulary Setting:} Inspired from Detic, we define our work under the weakly-supervised open-vocabulary setting as it uses image-level labels for expanding the detector's vocabulary. However in this setting, the complete target vocabulary set is unknown, \ie, only a selected number of novel and base categories are used for ILS from ImageNet-21K in LVIS. To evaluate our model in an extensive open-vocabulary setting, we modify our ILS by considering a larger vocabulary. Specifically, we expand the vocabulary to five times its size in \cite{zhou2022detecting}, by applying ILS from randomly sampled 5K categories from ImageNet-21k, in addition to the LVIS base classes. Table~\ref{tab:strict_ovd} compares our strict OVD setting results with ViLD where our performance slightly degrades showing sensitivity to ILS. However, we expect a gain with longer training as in Table~\ref{tab:lvis_compariosn}. In addition to above two settings, we train our LVIS model under stricter OVD conditions in a \textit{non} weakly-supervised setting by only using LVIS base categories for ILS. We achieve an overall 21.71 AP which is close to the model trained using ILS from 997 categories (22.75 AP). 

\begin{wraptable}{r}{6.3cm}
\vspace{-0.1in}
\setlength{\tabcolsep}{3pt}
\resizebox{1.0\linewidth}{!}{
  \begin{tabular}{lccc}
  \toprule
 Method & COCO & OpenImages & Objects365 \\
  \midrule
ViLD-text & 43.4 & - & 11.1 \\
Detic-base\dag  & 55.3 & 37.4 & 19.2 \\
\midrule
ViLD  & 55.6 & - & 18.2 \\
Detic\dag  & 56.3 & 42.2 & 21.7 \\
\midrule
Ours & \textbf{56.6} & \textbf{42.9} & \textbf{22.3} \\						
  \bottomrule
\end{tabular}
}
\caption{Cross-dataset evaluation. \dag The results evaluated using official implementation.}
\label{tab:ablation_crossds}
\vspace{-0.2in}
\end{wraptable}
\textbf{Cross-dataset evaluation performance:}
We provide cross-dataset evaluation of our model in Table~\ref{tab:ablation_crossds} and compare with prior OVD works. ViLD-text\cite{gu2021open} and Detic-base\cite{zhou2022detecting} are box-supervised baseline models for ViLD and Detic respectively. Our method builds on top of Detic-base and shows favourable results when directly transferred to cross-datasets without any dataset-specific finetuning. We use our method trained on LVIS and report AP$_{50}$ on COCO~\cite{coco}, OpenImages~\cite{kuznetsova2020open} and Objects365~\cite{shao2019objects365}.

\subsection{Analysis of RKD and ILS}
\textbf{Effect of Region-based Knowledge Distillation (RKD):} We ablate the effect of $\mathcal{L}_{1}$ (Eq.~\ref{l1_loss}) and $\mathcal{L}_{irm}$ (Eq.~\ref{l_IRM}) RKD approach on COCO (Table~\ref{tab:ablation_RKD}). The results show the importance of both loss functions, where using $\mathcal{L}_{1}$ loss over base model with top-5 proposals from MViT~\cite{maaz2021multi} improves the base and novel class by 1.9 and 15.0 AP (row-1 vs 3). Using $\mathcal{L}_{irm}$ in row-4 further improves the overall and novel class AP. To show the importance of using quality proposals in RKD, we compare the model trained with $\mathcal{L}_1$ loss using top-5 RPN vs MViT proposals (row-2 vs 3). All the models in rows 2-4 are finetuned on the base model.
\begin{table*}[h]
\small
\setlength\tabcolsep{4.4pt}
\begin{minipage}[t]{.45\linewidth}
\centering
\begin{tabular}{lccccc}
\toprule
\multirow{1}*{Method} & \multirow{1}*{AP$_{\txt{novel}}$} & \multirow{1}*{AP$_{\txt{base}}$} & \multirow{1}*{AP} \\
\midrule
1: Supervised (Base) & 1.7	& 53.2	& 39.6 \\
\midrule
2: RPN proposals $\mathcal{L}_1$ loss & 4.0 & {54.9} & 41.6 \\
3: MViT prop - $\mathcal{L}_1$ loss  & 16.7 & \textbf{55.1} & 45.0 \\
\rowcolor{orange!6}4: $\mathcal{L}_1$ + IRM loss & \textbf{21.2} & 54.7 & \textbf{45.9} \\ \bottomrule
\end{tabular}
\caption{Analysis on our region-based KD.}
\label{tab:ablation_RKD}
\end{minipage}
\hfill
\begin{minipage}[t]{.45\linewidth}
\centering
\setlength\tabcolsep{4.2pt}
\begin{tabular}{lccccc}
\toprule
\multirow{1}*{Method} & \multirow{1}*{AP$_{\txt{novel}}$} & \multirow{1}*{AP$_{\txt{base}}$} & \multirow{1}*{AP} \\
\midrule
1: Supervised (Base) & 1.7	& \textbf{53.2}	& 39.6 \\
\midrule
2: Max-Score loss on RPN & 15.9 & 48.2 & 39.7 \\
3: Max-Size loss on RPN & 25.9 & 51.1 & 44.5 \\
4: Max-Size of MViT & 28.9 & 50.7 & 45.0 \\
\rowcolor{orange!6}5: Pseudo-box on MViT & \textbf{30.4} & 52.6 & \textbf{46.8} \\ \bottomrule
\end{tabular}
\caption{Analysis on our weak IL supervision.}
\label{tab:ablation_ils}
\end{minipage}
\vspace{-0.1cm}
\end{table*}

\textbf{Effect of Weak Image-level Supervision (ILS):} We compare different choices of ILS in Table~\ref{tab:ablation_ils}. Our $\mathcal{L}_{pms}$ loss (Eq.~\ref{l_pms}) is compared with previously adopted ILS approaches~\cite{redmon2017yolo9000, ramanathan2020dlwl, zhou2022detecting} (rows 2-3). In row-4, we generate class-agnostic object proposals using ‘\txt{all objects}' text query with multi-modal ViTs (MViTs)~\cite{maaz2021multi} and select max-size proposal for ILS. In row-5, our proposed ILS approach uses target specific ‘\txt{every \{category\}}' text query with MViT and selects top-1 proposal for each ILS category. Our method (row-5) shows better performance compared to other alternatives. Additionally, we present all ablations on LVIS dataset in Appendix~\ref{appendix:ablations}.

\section{Qualitative Results}\label{appendix:qualitative_results}
\vspace{-1em}
\begin{figure*}[h]
\centering
{\includegraphics[width=0.83\textwidth]{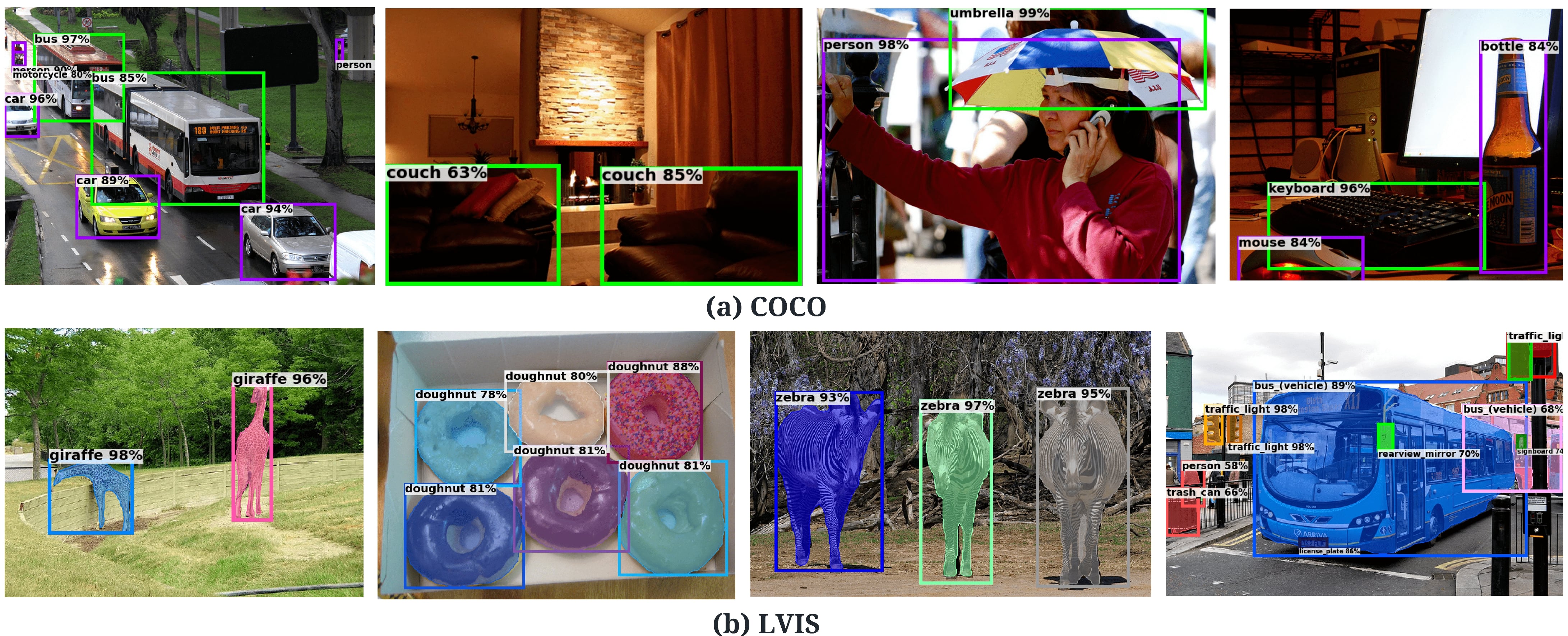}}
\caption{Qualitative results on \textbf{({\color{blue}{a}})} COCO and \textbf{({\color{blue}{b}})} LVIS images. For COCO, base and novel categories are shown in {\color{purple}{purple}} and {\color{green}{green}} colors respectively.
}
\label{fig:supplementary_fig_1}
\vspace{+0.1in}
\end{figure*}
\vspace{-2em}
\begin{figure*}[h]
\centering
{\includegraphics[width=0.83\textwidth]{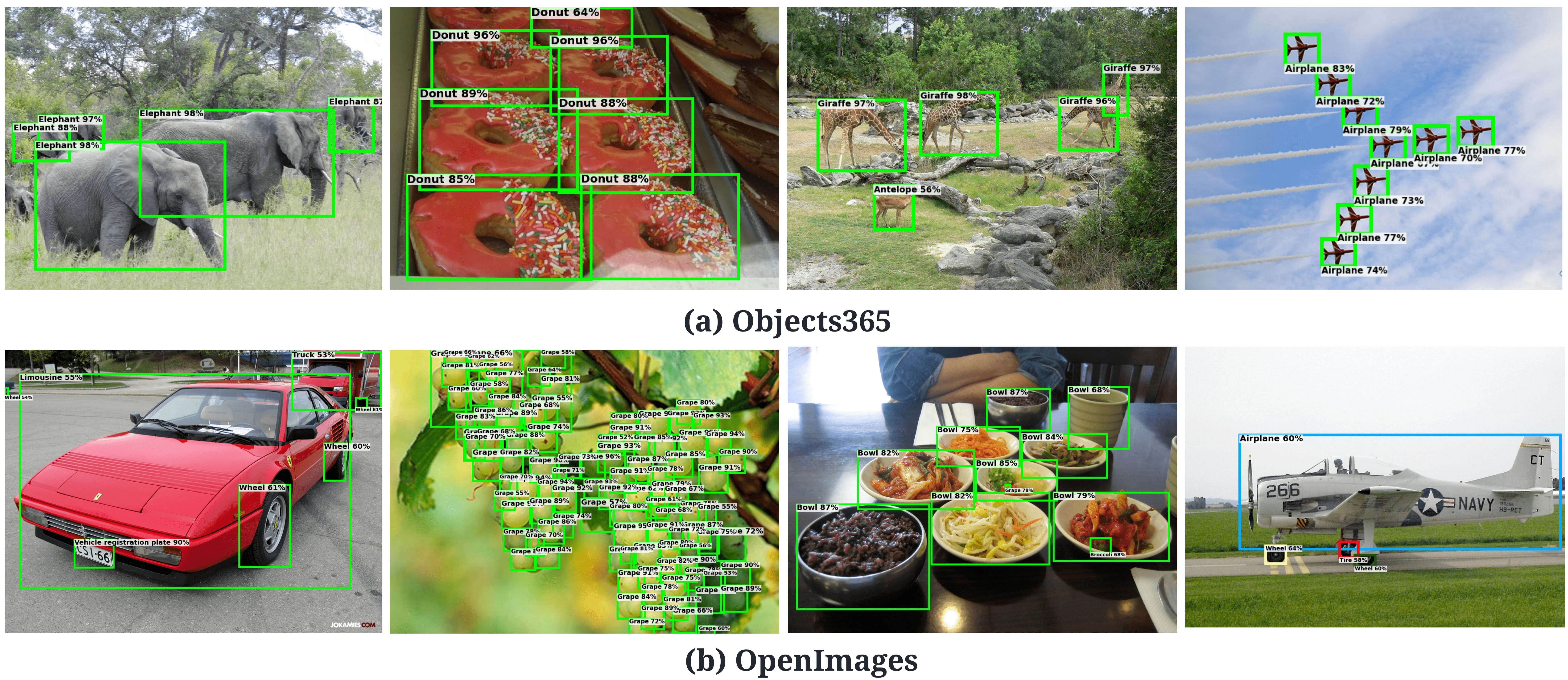}}
\caption{Qualitative results of cross-dataset transfer of our LVIS OVD model on \textbf{({\color{blue}{a}})} Objects365 and \textbf{({\color{blue}{b}})} OpenImages. Without any finetuning, our method provides high-quality detections.}
\label{fig:supplementary_fig_2}
\vspace{-0.2in}
\end{figure*}

\section{Conclusion}
This paper develops a novel framework to leverage the representation and generalization capability of pre-trained multi-modal models towards improved open-vocabulary detection (OVD). Specifically, we note that the existing OVD methods use weak supervision modes that are more image-centric, rather than object-centric for the end detection task. We proposed a novel knowledge distillation approach together with object-level pseudo-labeling to promote region-wise alignment between visual and language representations. Our weight transfer module provide an integration mechanism to combine the benefits of knowledge distillation and object-level pseudo-labeling. We demonstrate encouraging results on four popular OVD benchmarks, demonstrating sound generalization ability.

\textbf{Acknowledgements:}
The computations were performed in the Berzelius resource provided by the Knut and Alice Wallenberg Foundation at the National Supercomputer Centre.

{\small
\bibliographystyle{unsrt}
\bibliography{neurips_2022}
}

\section*{Checklist}

\begin{enumerate}

\item For all authors...
\begin{enumerate}
  \item Do the main claims made in the abstract and introduction accurately reflect the paper's contributions and scope?
    \answerYes{} The abstract and introduction clearly reflects the main contributions and scope of the paper.
  \item Did you describe the limitations of your work?
    \answerYes{} We have discussed the limitations of our work. Please refer to Appendix \ref{appendix:limitations}.
  \item Did you discuss any potential negative societal impacts of your work?
    \answerYes{} Please refer to Appendix \ref{appendix:negative_impacts}.
  \item Have you read the ethics review guidelines and ensured that your paper conforms to them?
    \answerYes{} We have read the ethics review guidelines and discussed the ethical implications of our work in Appendix \ref{appendix:ethical}.
\end{enumerate}

\item If you are including theoretical results...
\begin{enumerate}
  \item Did you state the full set of assumptions of all theoretical results?
    \answerNA{} Our findings and propositions are mainly based on experiments and empirical results. However, we have added relevant mathematical information in our theoretical formulations.
        \item Did you include complete proofs of all theoretical results?
    \answerNA{}
\end{enumerate}

\item If you ran experiments...
\begin{enumerate}
  \item Did you include the code, data, and instructions needed to reproduce the main experimental results (either in the supplemental material or as a URL)?
    \answerYes{}{} We provide the code along with the instructions to reproduce our main experiments in the supplemental material.
  \item Did you specify all the training details (e.g., data splits, hyperparameters, how they were chosen)?
    \answerYes{} We have provided all the training details including the data splits and hyperparameter choices in our paper. Please refer to sections \ref{checklist_dataset_details} and  \ref{checklist_implementation_details}.
        \item Did you report error bars (e.g., with respect to the random seed after running experiments multiple times)?
    \answerNo{} Due to the limited availability of compute resources, we have not reported these statistics.
        \item Did you include the total amount of compute and the type of resources used (e.g., type of GPUs, internal cluster, or cloud provider)?
    \answerYes{} Yes we have provided these details in the main paper. Please refer to the section \ref{checklist_implementation_details}.
\end{enumerate}

\item If you are using existing assets (e.g., code, data, models) or curating/releasing new assets...
\begin{enumerate}
  \item If your work uses existing assets, did you cite the creators?
    \answerYes{} We have cited all relevent existing works and assets which are related/used in our work.
  \item Did you mention the license of the assets?
    \answerYes{} We provide license details of the assets used in our work. Please refer to section \ref{checklist:asset_license}.
  \item Did you include any new assets either in the supplemental material or as a URL?
    \answerYes{} We provide our code for reproducing main experiments of our work in the supplemental material.
  \item Did you discuss whether and how consent was obtained from people whose data you're using/curating?
    \answerNA{} We use publically available datasets for our experiments. We have not explicitly discussed such consent in the main paper, but we have checked and made sure that all used datasets are allowed to be used for research.
  \item Did you discuss whether the data you are using/curating contains personally identifiable information or offensive content?
     \answerYes{} We discuss this in the supplemental material.
\end{enumerate}

\item If you used crowdsourcing or conducted research with human subjects...
\begin{enumerate}
  \item Did you include the full text of instructions given to participants and screenshots, if applicable?
    \answerNA{}
  \item Did you describe any potential participant risks, with links to Institutional Review Board (IRB) approvals, if applicable?
    \answerNA{}
  \item Did you include the estimated hourly wage paid to participants and the total amount spent on participant compensation?
    \answerNA{}
\end{enumerate}

\end{enumerate}

\newpage


\appendix
\begin{center}
\textbf{\Large Supplemental Material}
\end{center}

In this section, we provide additional information regarding,
\begin{itemize}
    \item Implementation details (Appendix~\ref{appendix:impl_details})
    \item Qualitative Results (Appendix~\ref{appendix:qualitative_results})
    \item Zero-shot Region Classification (Appendix~\ref{appendix_zero_shot})
    \item Additional Ablation Experiments (Appendix~\ref{appendix:ablations})
    \item Pseudo-labeling using Multi-modal ViTs (Appendix~\ref{appendix:mvit})
    \item Limitations (Appendix~\ref{appendix:limitations})
    \item Potential Negative Social Impacts (Appendix~\ref{appendix:negative_impacts})
    \item Ethical Considerations (Appendix~\ref{appendix:ethical})
    \item Datasets License Details (Appendix~\ref{checklist:asset_license})
\end{itemize}

\section{Implementation Details}
\label{appendix:impl_details}

We provide additional implementation details for our approach and datasets used in this work. We use standard Faster R-CNN~\cite{ren2015faster} with ResNet-50 C4 backbone and Mask R-CNN~\cite{he2017mask} with ResNet-50 FPN backbone for COCO and LVIS experiments respectively. We use L2 normalization on the region and text embeddings before computing the RKD loss and final classification scores. We note that this normalization is helpful to stabilize the training. For ILS, we sample images from detection and classification datasets with a ratio of 1:4. Specifically, we use a batch size of 16 and 64 for detection and classification datasets, respectively. We will release our codes and pretrained models publicly to ensure reproducibility of our results. 

\textbf{Datasets for weak Image-level Supervision (ILS):}
We use COCO captions and ImageNet-21k~\cite{deng2009imagenet} datasets for our proposed Image Level supervision (ILS) on COCO and LVIS datasets respectively. COCO captions dataset uses images from COCO detection dataset and provides five captions for each image. 
The words in a caption are compared heuristically, with every category name in the list of categories in COCO (base + novel). Using this method, we generate a list of positive categories for each image which is used as labels for ILS. We use ImageNet-21k \cite{russakovsky2015imagenet} for LVIS experiments which is a large scale classification dataset containing approximately 14M images and 21K classes. We use categories from ImageNet-21k which overlaps with LVIS categories, resulting in a subset containing 997 categories.

\textbf{Cross-dataset evaluation:} We provide cross-dataset evaluation of our LVIS trained model in Table~\ref{tab:ablation_crossds}. Following~\cite{zhou2022detecting,gu2021open}, we use validation sets of OpenImages V5 containing $\sim$41K images and Objects365 V2 containing $\sim$80K images for evaluation. We report AP$_{50}$ for cross-data evaluation. 

\section{Zero-shot Region Classification}
\label{appendix_zero_shot}

We compare the zero-shot classification performance of open-vocabulary detector with pretrained CLIP~\cite{radford2021learning} model on COCO validation dataset. Table~\ref{tab:clip_bs} shows the results where the top-1 classification accuracy is evaluated using the ground-truth object bounding boxes from COCO. The CLIP pretrained model shows better results for novel classes as compared to supervised-base model, indicating the strong generalization of the CLIP (row-1 vs 2). However the base class accuracy is higher for the supervised-base model as it is trained using COCO base classes. Further, using our region-based knowledge distillation (RKD) and novel weight transfer function improves the base and novel class performance, indicating the object-centric alignment in latent space. 

\begin{table}[h]
  \small
  \centering
  \setlength\tabcolsep{5pt}
  \begin{tabular}{lccc}
  \toprule
 \multirow{1}*{Method} & Top-1$_{base}$ & Top-1$_{novel}$ & Top-1$_{overall}$ \\
\midrule
1: Supervised (Base) & 88.8 & 42.5 & {76.7}\\
\midrule
2: CLIP & {57.3} & {59.4} & {57.8} \\
3: RKD  & {86.0} & {60.2} & {79.2} \\
\rowcolor{orange!6}4: Weight transfer  & {90.3} & {82.2} & {88.2} \\
\bottomrule
\end{tabular}
\vspace{0.15cm}
\caption{Classification results on novel and base classes with boxes cropped from COCO validation dataset using ground truth annotations. The pretrained CLIP shows competitive novel class accuracy. Our proposed RKD and weight transfer approach further improve the performance.}
\label{tab:clip_bs}
\vspace{-0.35cm}
\end{table}

\section{Additional Ablation Experiments}\label{appendix:ablations}

\subsection{Ablation Experiments on LVIS}

\textbf{Effect of individual components:} Table~\ref{tab:our_approach_summary_lvis} shows the contribution of individual components in our proposed approach on LVIS dataset. The baseline Mask-RCNN model (row-1) is trained on LVIS frequent and common classes using only the box-level supervision along with the zero-shot CLIP~\cite{radford2021learning} classifier. The results indicate the effectiveness of our region-based distillation (RKD) which explicitly aligns the image-centric CLIP embeddings to the object-centric region embeddings. Our image-level supervision (ILS) which uses class-specific pseudo-labels from the pretrained multi-modal ViT~\cite{maaz2021multi}, effectively enlarges the detector's vocabulary indicated by an increase of 4.8 AP over the base model for rare categories. Further, our proposed weight transfer scheme combines the strengths of the two methods and achieves better results on the common and frequent categories, while performing on par for the rare classes compared to naively combining the two approaches (row-4 vs 5).

\begin{table}[h!]
\hspace{-0.6em}
\centering
\begin{tabular}{lcccccc}
\toprule
\multirow{1}*{Method} & \multirow{1}*{AP$_{\txt{r}}$} & \multirow{1}*{AP$_{\txt{c}}$} & \multirow{1}*{AP$_{\txt{f}}$} & \multirow{1}*{AP} \\
\midrule
1: Supervised (Base) & 12.2	& 19.4	& 26.4 & 20.9 \\
\midrule
2: Base + Region based ditillation (RKD) & 15.2 & 20.2 & 27.3 & 22.1 \\
3: Base + ILS with pseudo-box (PIS) & 17.0 & 21.2 & 26.1 & 22.4 \\
4: RKD + PIS & 17.3 & 20.9 & 25.5 & 22.1 \\
\rowcolor{orange!6} 5: RKD + PIS + Weight-transfer (Ours) & 17.1 & 21.4 & 26.7 & 22.8 \\ \bottomrule
\end{tabular}
\vspace{0.15cm}
\caption{Effect of individual components in our method on LVIS dataset. Using RKD provides improvements over the baseline in all metrics (row-1 vs 2). Using ILS mainly helps in improving rare class performance (row-1 vs 3). Simply combining two methods shows improvements over the baseline but struggles to retain the individual performances especially for common and frequent categories (row-4). Our weight transfer approach provides complimentary gains from RKD and ILS, achieving good results as compared to simply adding both components (row-4 vs 5).}
\label{tab:our_approach_summary_lvis}
\vspace{0.2cm}
\end{table}

\textbf{Effect of Region-based Knowledge Distillation (RKD):} Table~\ref{tab:ablation_RKD_lvis} shows the effect of different loss functions ($\mathcal{L}_{1}$ and $\mathcal{L}_{irm}$ in Eq.~\ref{l1_loss} and Eq.~\ref{l_IRM} respectively) used in our region-based knowledge distillation (RKD) on LVIS dataset. It shows the effectiveness of using proposals from multi-modal ViT (MViT)~\cite{maaz2021multi} as compared to RPN for region-level alignment (row-2 vs 3). Using high-quality MViT proposals provides significant gains compared to using RPN proposals. Further, using our inter-embedding relationship matching (IRM) loss along with $\mathcal{L}_{1}$ loss provides an overall good trade-off between rare, common and frequent class AP.

\begin{table}[h!]
\centering

\begin{tabular}{lcccccc}
\toprule
\multirow{1}*{Method} & \multirow{1}*{AP$_{\txt{r}}$} & \multirow{1}*{AP$_{\txt{c}}$} & \multirow{1}*{AP$_{\txt{f}}$} & \multirow{1}*{AP} \\
\midrule
1: Supervised (Base) & 12.2	& 19.4	& 26.4 & 20.9 \\
\midrule
2: RPN proposals $\mathcal{L}_1$ loss & 8.7 & 17.4 & 26.1 & 19.3 \\
3: MViT prop - $\mathcal{L}_1$ loss  & 12.4 & 20.7 & 27.7 & 22.0 \\
\rowcolor{orange!6}4: $\mathcal{L}_1$ + IRM loss & 15.2 & 20.2 & 27.3 & 22.1 \\ \bottomrule
\end{tabular}

\vspace{0.15cm}
\caption{Analysis on our RKD method on LVIS.}
\label{tab:ablation_RKD_lvis}
\end{table}

\textbf{Effect of Weak Image-level Supervision (ILS):} Table~\ref{tab:ablation_ils_lvis} compares the different heuristics based approaches opted for image-level supervision (ILS) versus our method that utilizes class-specific proposals from the pretrained MViT on LVIS dataset. Selecting top-1 proposal from MViT using target specific specific queries such as ‘\txt{every \{category\}}' provides optimal performance for rare classes.

\begin{table}[h!]
\centering

\begin{tabular}{lcccccc}
\toprule
\multirow{1}*{Method} & \multirow{1}*{AP$_{\txt{r}}$} & \multirow{1}*{AP$_{\txt{c}}$} & \multirow{1}*{AP$_{\txt{f}}$} & \multirow{1}*{AP} \\
\midrule
1: Supervised (Base) & 12.2	& 19.4	& 26.4 & 20.9 \\
\midrule
2: Max-Score loss on RPN & 12.8 & 18.6 & 24.7 & 20.0 \\
3: Max-Size loss on RPN & 14.9 & 21.3 & 26.1 & 22.1 \\
\rowcolor{orange!6}4: Pseudo-box on MViT & 17.0 & 21.2 & 26.1 & 22.4 \\ 
\bottomrule
\end{tabular}

\vspace{0.15cm}
\caption{Analysis on our weak ILS on LVIS.}
\label{tab:ablation_ils_lvis}
\end{table}


\subsection{Initialization for RKD Training}

We note that it is important to properly initialize the RKD training to gain its full advantages. Table~\ref{tab:initialization_rkd} shows that training RKD from scratch (row-2) results in lower base class AP. However, initializing the RKD training from the Supervised base model recovers this loss and provides improvements over the base model. This indicates that region-based alignment is sensitive to the distribution of the features and requires mature features for effectively distilling knowledge from pretrained CLIP model. This observation is same as in~\cite{joseph2021towards} where the contrastive clustering is enabled only on the mature features after a few training epochs for open-world object detection.

\begin{table}[h!]
\centering

\begin{tabular}{lccccc}
\toprule
\multirow{1}*{Method} & \multirow{1}*{AP$_{\txt{novel}}$} & \multirow{1}*{AP$_{\txt{base}}$} & \multirow{1}*{AP} \\
\midrule
1: Supervised (Base) & 1.7	& 53.2	& 39.6 \\
\midrule
2: RKD from scratch & 21.3 & 50.9 & 43.1 \\
\rowcolor{orange!6}3: Base + RKD  & 21.2 & 54.7 & 45.9 \\
\bottomrule
\end{tabular}

\vspace{0.15cm}
\caption{Effect of initialization for RKD training on COCO dataset.}
\label{tab:initialization_rkd}
\end{table}



\subsection{Additional Ablation Experiment}
Table~\ref{tab:ablation_mlp_skip_conn} shows the ablation on using a MLP skip connection across $\mathcal{W_P}$ in Fig.~\ref{ovd_block_diag}. We add this skip connection to form a direct path for region classification using CLIP in ILS. This allows the weight transfer function to specifically focus on the residual signal in the ILS pathway. It improves the convergence and helps to attain better results in most cases on LVIS/COCO datasets.

\begin{table}[h!]
\centering
\resizebox{0.87\linewidth}{!}{
\begin{tabular}{l|ccc|cccc}
\toprule
& \multicolumn{3}{c|}{COCO} & \multicolumn{4}{c}{LVIS} \\
\midrule
\multirow{1}*{Method} & \multirow{1}*{AP$_{\txt{novel}}$} & \multirow{1}*{AP$_{\txt{base}}$} & \multirow{1}*{AP} & \multirow{1}*{AP$_r$} &
\multirow{1}*{AP$_c$} & \multirow{1}*{AP$_f$} & \multirow{1}*{AP}\\
\midrule
1: Supervised (Base) & 1.7	& 53.2	& 39.6 & 12.2 & 19.4 & 26.4 & 20.9 \\
\midrule
\rowcolor{orange!6}2: RKD + PIS + Weight-transfer (Ours) & 36.6 & 54.0 & 49.4 & 17.1 & 21.4 & 26.7 & 22.8 \\
3: + w/o MLP skip connection  & 32.5 & 53.5 & 48.0 & 18.1 & 20.9 & 26.2 & 22.5 \\
\bottomrule
\end{tabular}
}
\vspace{0.15cm}
\caption{The ablation on using MLP skip connection in Fig.~\ref{ovd_block_diag}.}
\label{tab:ablation_mlp_skip_conn}
\end{table}

\section{Pseudo Labeling using Multi-modal ViTs}
\label{appendix:mvit}
\begin{figure*}[!t]
\centering
{\includegraphics[width=1.0\textwidth]{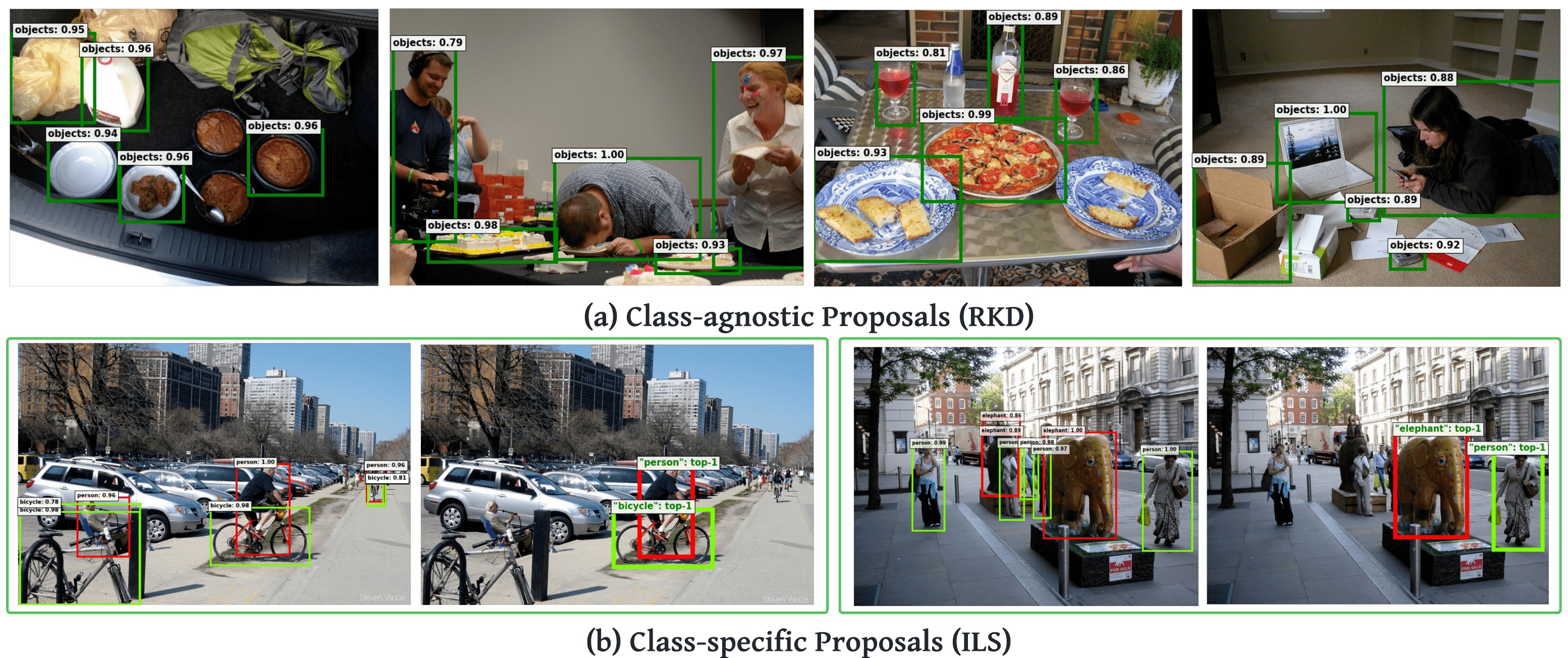}}
\caption{\small \textbf{\color{blue}{(a)}} \textbf{Class-agnostic Proposals:} The figure shows the top 5 class-agnostic proposals obtained from the MViT~\cite{maaz2021multi} using ‘\txt{all objects}' text query. As illustrated, these high-quality tightly bound object proposals provide rich local-semantics for RKD in our proposed pipeline. \textbf{\color{blue}{(b)} } \textbf{Class-specific Proposals:} The figure shows the class-specific proposals obtained from the MViT using ‘\txt{every <category name>}' text queries. The left image in each pair shows all proposals while the corresponding right image shows the selected top 1 proposal per category for ILS.}
\label{fig:ablation:mavl}
\end{figure*}

In this section, we describe the process of generating class-agnostic and class-specific proposals using multi-modal ViTs (MViTs)~\cite{maaz2021multi,kamath2021mdetr}. We name this process as \emph{pseudo labeling $\mathcal{Q}_{\text{pseudo}}$}. The MViT model is trained using aligned image text pairs and is capable of locating novel and base class objects using relevant human-intuitive text queries. For example, targeted text queries such as ‘\txt{every person}' and ‘\txt{every elephant}' can be used to locate all persons and all elephants in an image respectively (Fig.~\ref{fig:ablation:mavl}b). Maaz~\etal~\cite{maaz2021multi} show that the MViTs encode the object-centric concepts using aligned image-caption pairs and are excellent class-agnostic object detectors. The authors designed text queries such as ‘\txt{all objects}' and ‘\txt{all entities}' and demonstrated state-of-the-art class-agnostic object detection results on multiple datasets across different domains. We use these MViTs to generate class-agnostic and class-specific object proposals for region-based knowledge distillation (RKD) and weak image-level supervision (ILS), respectively.

\textbf{Class-agnostic proposals for RKD:} We generate class-agnostic object proposals from the MViT~\cite{maaz2021multi} using ‘\txt{all objects}' text query. The generated proposals are ranked using predicted objectness scores and the top 5 proposals per image are selected for RKD as shown in Fig.~\ref{fig:ablation:mavl}a. Next, the CLIP~\cite{radford2021learning} image-encoder and our OVD detector is used to generate embeddings corresponding to these proposals which are then used for calculating the RKD loss in Eq.~\ref{RKD_loss}. To save the computation load and increase the training efficiency, we compute the class-agnostic proposals and the corresponding CLIP region embeddings offline and load them during training. Further for LVIS experiments, we use images from a subset of ImageNet-21K (consisting of 997 overlapping LVIS categories) for RKD as well.

\textbf{Class-specific proposals for ILS:} We generate class-specific proposals from the MViT~\cite{maaz2021multi} using ‘\txt{every <category name>}' text query. Given the $N$ category names present in an image, we use $N$ queries of format ‘\txt{every <category name>}' to generate class-specific proposals followed by selecting top 1 proposal for each category. This provides us $N$ high-quality box proposals per image corresponding to $N$ categories present in the image. 
These proposals are used to effectively enhance the detector's vocabulary using ILS during training. Further, to maintain the training efficiency of our experiments, we compute these class-specific proposals offline and load them during training.

\section{Limitations} \label{appendix:limitations}
Our proposed OVD method encourages object centric visual-language (VL) alignment using a novel weight transfer method which combines benefits from RKD and ILS. Irrespective of the state-of-the-art results on novel/rare classes, there is still a significant gap between base and novel class performances (e.g. 56.7 and 40.5 AP for COCO base and novel categories in Table~\ref{tab:coco_comparison}, 29.1 and 21.1 Mask AP for LVIS frequent and rare categories in Table~\ref{tab:lvis_compariosn}). 
Further, the open-vocabulary capabilities of our model largely depends or are limited to the vocabulary of the pretrained CLIP~\cite{radford2021learning} model, which is used as a teacher in our RKD pipeline.


\section{Potential Negative Social Impacts} \label{appendix:negative_impacts}
The results of cross-dataset transfer evaluations show that the vocabulary of our detector is highly flexible and can be expanded to any number of categories, based on the downstream tasks and datasets. This poses a risk on how our OVD detector with a large vocabulary can be used in inappropriate ways in the community such as for large scale illegal video surveillance. Furthermore, OVD capabilities can be modulated for targeted detections instead of generic detections by tuning the classifier weights using specialized prompts. This could add biases in the detector and can lead to unfair predictions.

\section{Ethical Considerations}\label{appendix:ethical}
The OVD response to recognize object categories strongly depends on the image-text pretraining datasets used for the training of VL model (CLIP in our case). Thus, the source of these datasets can pose ethical issues. For example, datasets extracted from internet can contain racial and unethical bias and can modulate the ethical behaviour of the detector as well. Thus, before applying our OVD detector in a practical scenario, such biases of the pretraining/training datasets should be removed to have fairness and ethically correct results of the detector.
Moreover, the detector vocabulary is flexible and it can be tuned to show racial biasness while detecting humans. For example, weights of the zero-shot classifier generated with specialized biased prompts could lead to biased and unethically targeted human detections (e.g., black vs white) which must be taken into consideration.

\section{License Details}\label{checklist:asset_license}
\begin{wraptable}[19]{r}{6.7cm}
\vspace{-0.15in}
\setlength{\tabcolsep}{3pt}
\resizebox{1.0\linewidth}{!}{
\begin{tabular}{lcccccc}
\toprule
\multirow{1}*{Dataset}  & Task & \multirow{1}*{License} \\
\midrule
COCO & OVD  &  Custom (CC BY 4.0) \\
\midrule
\multirow{2}*{LVIS v1.0}  & \multirow{2}*{OVD} &  \multirow{1}*{CC BY 4.0 \&}\\
 &  &  \multirow{1}*{ COCO license}\\
\midrule
{ImageNet-21K}  & ILS in LVIS & CC BY-NC\\
\midrule
\multirow{1}*{Flickr30k}  & \multirow{1}* {MViT} & \multirow{1}* {CC BY-NC}\\
\midrule
\multirow{1}*{Visual Genome} & {MViT} &\multirow{1}* {CC BY 4.0}\\
\midrule
\multirow{1}*{GQA} & MViT   & \multirow{1}* {CC BY 4.0}\\
\midrule
\multirow{2}*{Objects365} &  \multirow{1}* {Cross-data} & \multirow{2}* {CC BY 4.0}\\
 &  \multirow{1}* {evaluation} &\\
\midrule
\multirow{2}*{OpenImages}  & {Cross-data} &\multirow{2}* {CC BY 4.0}\\
 &  \multirow{1}* {evaluation} & \\
\midrule
\multirow{1}*{OpenImages}  & \multirow{1}*{Cross-data} &\multirow{1}* {Google LLC}\\
\multirow{1}*{annotations}  &    \multirow{1}*{evaluation}   &\multirow{1}* { \& CC BY 2.0}\\
\bottomrule
\end{tabular}
}
\caption{Summary of licenses for datasets used in our experiments.}
\label{table:license}
\vspace{-0.2in}
\end{wraptable}

Here we provide license details of the datasets used in our work, summarized in Table \ref{table:license}. COCO is available for non-commercial use under the Creative Commons Attribution 4.0 (CC BY 4.0) license. LVIS is based on the COCO dataset, and it is licensed under both CC BY 4.0 and the COCO license. ImageNet-21k is a publically available dataset available for research and non-commercial use. It is licensed under Creative Commons (CC), and its type is "CC BY-NC". We use a pretrained MViT model for proposal generation, which is trained on LMDet (Large scale Modulated Detection dataset). It uses Flicker30k, Visual Genome, and GQA datasets. The license type of Flicker30k is CC BY-NC. Visual Genome and GQA both have the same license type CC BY 4.0. For cross-datasets evaluation, Objects365 and OpenImages are used, which are licensed under Creative Commons Attribution 4.0. Annotations of OpenImages are licensed by Google LLC under Creative Commons Attribution 2.0.

\end{document}